\documentclass[11pt,a4paper]{article}

\usepackage[hyperref]{acl2019}
\usepackage{times}
\usepackage{latexsym}
\usepackage{tabularx}
\usepackage{booktabs}
\usepackage{graphicx}
\usepackage{amsmath}
\usepackage{ mathrsfs }
\usepackage{amsmath}
\DeclareMathOperator*{\argmax}{arg\,max}

\usepackage{subcaption}
\newcommand{\cameraready}[1]{\textcolor{black}{#1}}

\usepackage{url}

\aclfinalcopy 

\title{Strategies for Structuring Story Generation}

\author{Angela Fan \\
  FAIR, Paris \\
  LORIA, Nancy \\
  \texttt{angelafan@fb.com} \\\And
  Mike Lewis \\
  FAIR, Seattle \\
  \texttt{mikelewis@fb.com} \\\And
  Yann Dauphin \\
  Google AI\footnotemark \\
  \texttt{ynd@google.com} \\}

\date{}

\begin{document}
\maketitle

\renewcommand*{\thefootnote}{\fnsymbol{footnote}}
\footnotetext{*Work done while at Facebook AI Research}
\renewcommand*{\thefootnote}{\arabic{footnote}}

\begin{abstract}
Writers often rely on plans or sketches to write long stories, but most current language models generate word by word from left to right.
We explore coarse-to-fine models for creating narrative texts of several hundred words, and introduce new models which decompose stories by abstracting over actions and entities.
The model first generates the predicate-argument structure of the text, where different mentions of the same entity are marked with placeholder tokens.
It then generates a surface realization of the predicate-argument structure, and finally replaces the entity placeholders with context-sensitive names and references.
Human judges prefer the stories from our models to a wide range of previous approaches to hierarchical text generation.
Extensive analysis shows that our methods can help improve the diversity and coherence of events and entities in generated stories.

\end{abstract}

\section{Introduction}
Stories exhibit structure at multiple levels. While existing language models can generate stories with good local coherence, they struggle to coalesce individual phrases into coherent plots or even maintain character consistency throughout a story. One reason for this failure is that classical language models generate the whole story at the word level, which makes it difficult to capture the high-level interactions between the plot points.

\begin{figure}
   \centering
   \includegraphics[width=0.48\textwidth]{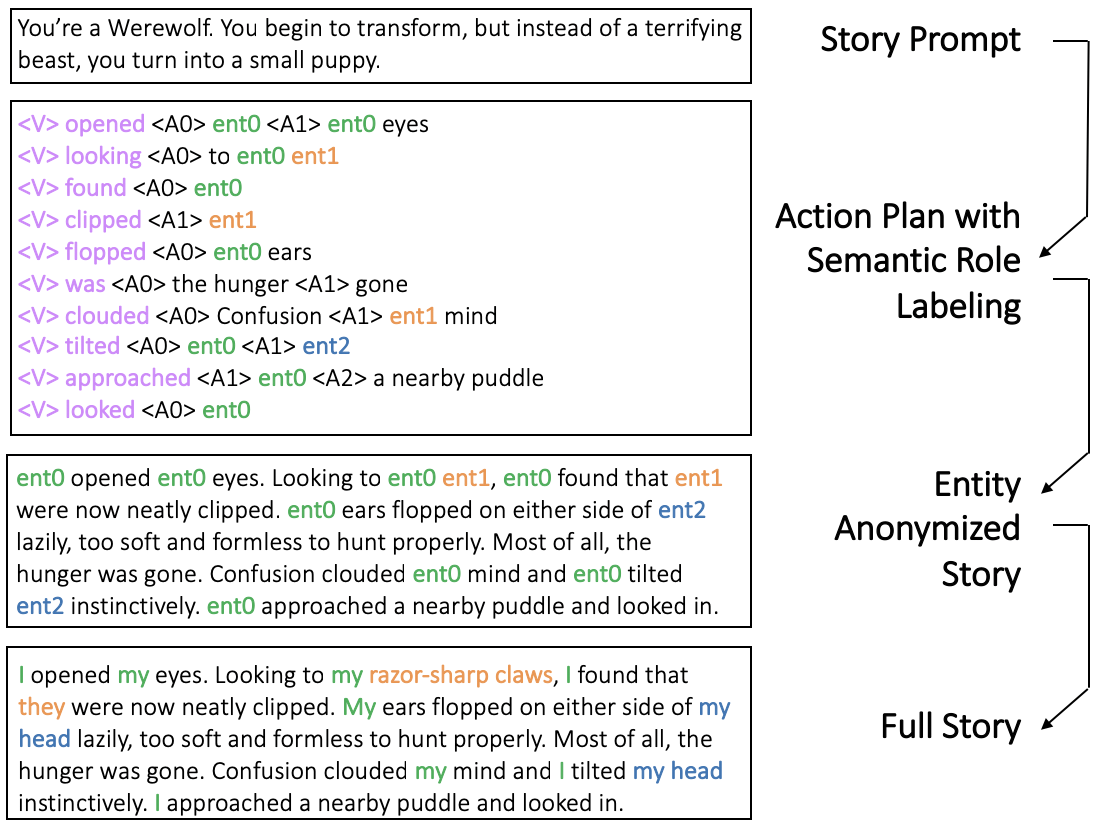}
 \caption{Proposed Model. Conditioned upon the prompt, we generate sequences of predicates and arguments. Then, a story is generated with placeholder entities such as \textit{ent0}. Finally we replace the placeholders with specific references.}
 \label{fig:entire_model}
\end{figure}

To address this, we investigate novel decompositions of the story generation process that break down the problem into a series of easier coarse-to-fine generation problems. These decompositions can offer three advantages:
\begin{itemize}
    \setlength\itemsep{-0.3em} 
    \item They allow more abstract representations to be generated first, where challenging long-range dependencies may be more apparent.
    \item They allow specialized modelling techniques for the different stages, which exploit the structure of the specific sub-problem.
    \item They are applicable to any textual dataset and require no manual labelling.
\end{itemize}

Several hierarchical models for story generation have recently been proposed \cite{xu2018skeleton, yao2019plan}, but it is not well understood which properties characterize a good decomposition. We therefore implement and evaluate several representative approaches based on keyword extraction, sentence compression, and summarization. 

We build on this understanding to devise the proposed decomposition (Figure \ref{fig:entire_model}). Inspired by the classic model of \newcite{reiter2000building}, our approach breaks down the generation process in three steps: modelling the action sequence, the \cameraready{story} narrative, and lastly entities such as story characters. To model action sequences, we first generate the predicate-argument structure of the story by generating a sequence of verbs and arguments. This representation is more structured than free text, making it easier for the model learn dependencies across events. To model entities, we initially generate a version of the story where different mentions of the same entity are replaced with placeholder tokens. Finally, we re-write these tokens into different references for the entity, based on both its previous mentions and the global story context. 

The models are trained on 300k stories from \textsc{WritingPrompts} \citep{fan2018hierarchical}, and we evaluate quality both in terms of human judgments and using automatic metrics. We find that our approach substantially improves story generation. Specifically, we show that generating the action sequence first makes the model less prone to generating generic events, leading to a much greater diversity of verbs. We also find that by using sub-word modelling for the entities, our model can produce novel names for locations and characters that are appropriate given the story context.

\section{Model Overview}

The crucial challenge of long story generation lies in maintaining coherence across a large number of generated sentences---in terms of both the logical flow of the story and the characters and entities. While there has been much recent progress in left-to-right text generation, particularly using self-attentive architectures \cite{dai2018transformer, liu2018generating}, we find that models still struggle to maintain coherence to produce interesting stories on par with human writing. We therefore introduce strategies to decompose neural story generation into coarse-to-fine steps to make modelling high-level dependencies easier to learn.

\subsection{Tractable Decompositions}
In general, we can decompose the generation process by converting a story $x$ into a more abstract representation $z$. The negative log likelihood of the decomposed problem is given by
\begin{align}
\mathcal{L} &= -\log \sum_z p(x|z)p(z).   \label{eq:marginalize}
\end{align}
We can generate from this model by first sampling from $p(z)$ and then sampling from $p(x|z)$. However, the marginalization over $z$ is in general intractable, except in special cases where every $x$ can only be generated by a single $z$ (for example, if the transformation removed all occurrences of certain tokens). Instead, we minimize a variational upper bound of the loss by constructing a deterministic posterior $q(z|x) = \mathrm{1}_{z=z^*}$, where $z^*$ can be given by running semantic role labeller or coreference resolution system on $x$. Put together, we optimize the following loss:
\begin{align}
z^* &= \argmax_z p(z|x) \\
\mathcal{L} &\leq -\log p(x|z^*) - \log p(z^*) \label{eq:decomposed}
\end{align}
This approach allows models $p(z^*)$ and $p(x|z^*)$ to be trained tractably and separately.

\subsection{Model Architectures}
We build upon the convolutional sequence-to-sequence architecture \cite{gehring2017convs2s}. Deep convolutional networks are used as both the encoder and decoder. The networks are connected with an attention module \cite{bahdanau2015neural} that performs a weighted sum of the encoder output. The decoder uses a gated multi-head self-attention mechanism \cite{vaswani2017transformer, fan2018hierarchical} to allow the model to refer to previously generated words and improve the ability to model long-range context.

\section{Modelling Action Sequences}
\begin{figure*}
   \centering
   \includegraphics[width=0.8\textwidth]{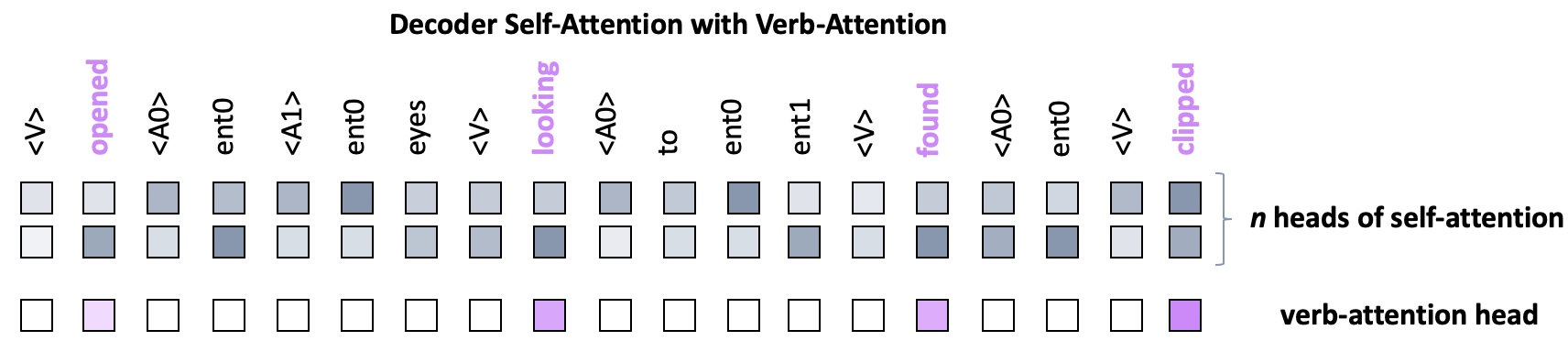}
 \caption{Verb-Attention. To improve the model's ability to condition upon past verbs, one head of the decoder's self-attention mechanism is specialized to only attend to previously generated verbs.}
 \label{fig:verb_attention}
\end{figure*}
To decompose a story into a structured form that emphasizes logical sequences of actions, we use Semantic Role Labeling (SRL). 
SRL identifies \emph{predicates} and \emph{arguments} in sentences, and assigns each argument a \emph{semantic role}.
This representation abstracts over different ways of expressing the same semantic content. For example, \emph{John ate the cake} and \emph{the cake that John ate} would receive identical semantic representations.

Conditioned upon the prompt, we generate an SRL decomposition of the story by concatenating the predicates and arguments identified by a pretrained model  \cite{he2017deep, tan2018deep}\footnote{for predicate identification, we use \url{https://github.com/luheng/deep\_srl}, for SRL given predicates, we use \url{https://github.com/XMUNLP/Tagger}} and separating sentences with delimiter tokens. We place the predicate verb first, followed by its arguments in canonical order. To focus on the main narrative, we retain only core arguments.

\paragraph{Verb Attention Mechanism} 
SRL parses are more structured than free text, enabling more structured models.
To encourage the model to consider sequences of verbs, we designate one of the heads of the decoder's multihead self-attention to be a \textit{verb-attention} head (see Figure~\ref{fig:verb_attention}). By masking the self-attention appropriately, this verb-attention head can only attend to previously generated verbs. When the text does not yet have a verb, the model attends to a zero vector. We show that focusing on verbs with a specific attention head generates a more diverse array of verbs and reduces repetition in generation.

\begin{figure*}
   \centering
   \includegraphics[width=\textwidth]{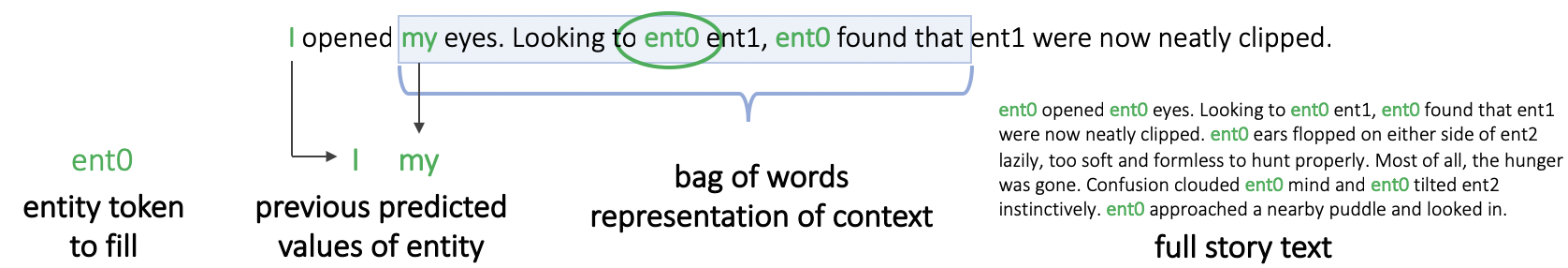}
 \caption{Input for Coreferent entity reference generation. The model has a representation of the entity context in a bag of words form, all previous predicted values for the same anonymized entity token, and the full text story. The green circle represents the entity mention the model is attempting to fill.}
 \label{fig:coref_fill}
\end{figure*}

\section{Modelling Entities}

The challenge of modelling characters throughout a story is twofold: first, entities such as character names are rare tokens, which make them hard to model for neural language models. Human stories often feature imaginative, novel character or location names. Second, maintaining the consistency of a specific set of characters is difficult, as the same entity may be referenced by many different strings throughout a story---for example \emph{Bilbo Baggins}, \emph{he}, and \emph{the hobbit} may refer to the same entity. It is challenging for existing language models to track which words refer to which entity purely using a language modelling objective.

We address both problems by first generating a form of the story with different mentions of the same entity replaced by a placeholder token (e.g. \emph{ent0}), similar to \citet{hermann2015teaching}. We then use a sub-word seq2seq model trained to replace each mention with a reference, based on its context. The sub-word model is better equipped to model rare words and the placeholder tokens make maintaining consistency easier.

\subsection{Generating Entity Anonymized Stories} 
We explore two approaches to identifying and clustering entities:
\begin{itemize}
    \setlength\itemsep{-0.3em}    
    \item \textbf{NER Entity Anonymization}: We use a named entity recognition (NER) model\footnote{\url{https://spacy.io/api/entityrecognizer}} to identify all people, organizations, and locations. We replace these spans with placeholder tokens (e.g. \emph{ent0}). If any two entity mentions have an identical string, we replace them with the same placeholder. For example, all mentions of \emph{Bilbo Baggins} will be abstracted to the same entity token, but \emph{Bilbo} would be a separate abstract entity.
    \item  \textbf{Coreference-based Entity Anonymization}: The above approach cannot detect different mentions of an entity that use different strings. Instead, we use the Coreference Resolution model from \citet{lee2018higher}\footnote{\url{https://github.com/kentonl/e2e-coref}} to identify clusters of mentions. All spans in the same cluster are then replaced with the same entity placeholder string. Coreference models do not detect singleton mentions, so we also replace non-coreferent named entities with unique placeholders.
\end{itemize}

\subsection{Generating Entity References in a Story} 

We train models to replace placeholder entity mentions with the correct surface form, for both NER-based and coreference-based entity anonymised stories. Both our models use a seq2seq architecture that generates an entity reference based on its placeholder and the story. To better model the specific challenges of entity generation, we also make use of a pointer mechanism and sub-word modelling.

\paragraph{Pointer Mechanism} 
Generating multiple consistent mentions of rare entity names is challenging. 
To aid re-use of previous names for an entity, we augment the standard seq2seq decoder with a pointer-copy mechanism \cite{vinyals2015pointer}. 
To generate an entity reference, the decoder can either generate a new abstract entity token or choose to copy an already generated abstract entity token, which encourages the model to use consistent naming for the entities. 

To train the pointer mechanism, the final hidden state of the model $h$ is used as input to a classifier $p_\text{copy}(h) = \sigma(w_\text{copy} \cdot h)$. $w_\text{copy}$ is a fixed dimension parameter vector. When the model classifier predicts to copy, the previously decoded abstract entity token with the maximum attention value is copied. One head of the decoder multi-head self-attention mechanism is used as the pointer copy attention head, to allow the heads to specialize. 

\paragraph{Sub-word Modelling}
Entities are often rare or novel words, so word-based vocabularies can be inadequate. We compare entity generation using word-based, byte-pair encoding (BPE) \cite{sennrich2015neural}, and character-level models. 

\paragraph{NER-based Entity Reference Generation} 
Here, each placeholder string should map onto one (possibly multiword) surface form---e.g. all occurrences of the placeholder \emph{ent0} should map only a single string, such as \emph{Bilbo Baggins}.
We train a simple model that maps a combination placeholder token and story (with anonymized entities) to the surface form of the placeholder. While the placeholder  can appear multiple times, we only make one prediction for each placeholder as they all correspond to the same string. 

\paragraph{Coreference-based Entity Reference Generation}
Generating entities based on coreference clusters is more challenging than for our NER entity clusters, because different mentions of the same entity may use different surface forms. 
We generate a separate reference for each mention by adding the following inputs to the above model:
\begin{itemize}
    \setlength\itemsep{-0.3em} 
    \item A \textit{bag-of-words} context window around the specific entity mention, which allows local context to determine if an entity should be a name, pronoun or nominal reference.
    \item \textit{Previously generated} references for the same entity placeholder. For example, if the model is filling in the third instance of \textit{ent0}, it receives that the previous two generations for \textit{ent0} were \textit{Bilbo, him}. Providing the previous entities allows the model to maintain greater consistency between generations.  
\end{itemize}

\section{Experimental Setup}

\subsection{Data}

We use the \textsc{WritingPrompts} dataset from \cite{fan2018hierarchical} \footnote{\url{https://github.com/pytorch/fairseq/tree/master/examples/stories}} of 300k story premises paired with long stories. Stories are on average 734 words, making the generation far longer compared to related work on storyline generation. In this work, we focus on the prompt to story generation aspect of this task. \cameraready{We assume models receive a human-written prompt, as shown in Figure~\ref{fig:entire_model}.} We follow the previous preprocessing of limiting stories to 1000 words and fixing the vocabulary size to 19,025 for prompts and 104,960 for stories.

\subsection{Baselines}

\begin{figure*}
   \centering
   \includegraphics[width=0.7\textwidth]{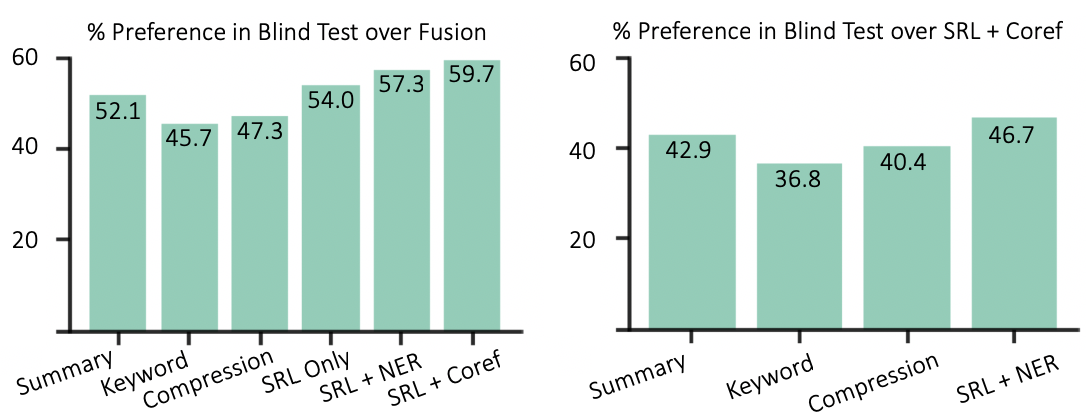}
 \caption{Human evaluations of different decomposed models for story generation. We find that using SRL action plans and coreference-resolution to build entity clusters generates stories that are preferred by human judges.}
 \label{fig:turk_studies}
\end{figure*}

We compare our results to the \textit{Fusion} model from \citet{fan2018hierarchical} which generates the full story directly from the prompt. We also implement various decomposition strategies as baselines:
\begin{itemize}
    \setlength\itemsep{-0.3em} 
    \item \textit{Summarization}: We propose a new baseline that generates a summary conditioned upon the prompt and then a story conditioned upon the summary. Story summaries are obtained with a multi-sentence summarization model \cite{wu2018pay} trained on \cameraready{the full-text version of the CNN-Dailymail summarization corpus \cite{hermann2015teaching,nallapati2016abstractive,see2017get}\footnote{\url{https://github.com/abisee/cnn-dailymail}}} and applied to stories.
    \item \textit{Keyword Extraction}: We generate a series of keywords conditioned upon the prompt and then a story conditioned upon the keywords, based on \citet{yao2019plan}. Following Yao et al, we extract keywords with the \textsc{rake} algorithm \cite{rose2010automatic}\footnote{\url{https://pypi.org/project/rake-nltk/}}. Yao et al. extract one word per sentence, but we find that extracting $n=10$ keyword phrases per story worked well, as our stories are much longer.
    \item \textit{Sentence Compression}: Inspired by \citet{xu2018skeleton}, we generate a story with compressed sentences conditioned upon the prompt and then a story conditioned upon the compression. We use the same deletion-based compression data as Xu et al., from \citet{filippova2013overcoming}\footnote{\url{https://github.com/google-research-datasets/sentence-compression}}. We train a seq2seq model to compress all non-dialog story sentences (as the training data does not contain much spoken dialogue). The compressed sentences are concatenated to form the compressed story. 
\end{itemize}

\begin{table}[t]
  \centering \small
  \begin{tabular}{ l p{13mm} p{15mm} }\hline
    \bf{Decomposition} & \bf{Stage 1} \newline $-\log p(z^*)$ & \bf{Stage 2} \newline  $-\log p(x|z^*)$\\ \hline\hline
    Summary & 4.20 & 5.09 \\
    Keyword & 6.92 & 4.23 \\ 
    Compression & 5.05 & 3.64 \\ 
    \hline 
    SRL Action Plan & 2.72 & 3.95 \\ 
    NER Entity Anonymization & 3.32 & 4.75 \\ 
    Coreference Anonymization & 3.15 & 4.55 \\ 
 \hline 
\end{tabular}
   \caption{Negative log likelihood of generating stories using different decompositions \cameraready{(lower is easier for the model)}. Stage 1 is the generation of the intermediate representation $z^*$, and Stage 2 is the generation of the story $x$ conditioned upon $z^*$. Entity generation is with a word-based vocabulary to be consistent with the other models.}
 \label{tbl:stage_losses}
\end{table}

\begin{figure}
   \centering
   \includegraphics[width=0.3\textwidth]{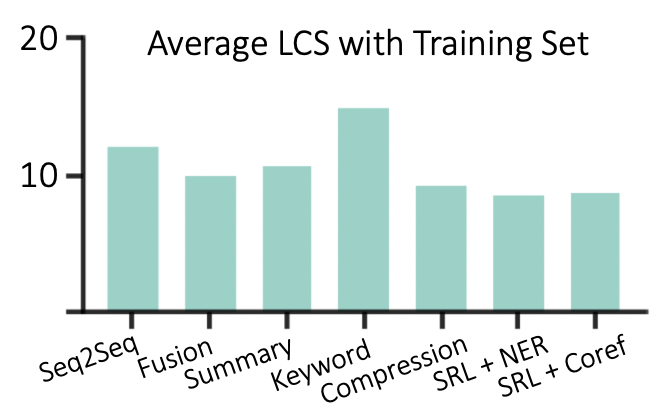}
 \caption{\cameraready{Average Longest Common Subsequence of Generated Stories with human-written stories in the training set.}}
 \label{fig:lcs}
\end{figure}

\subsection{Training}

We implement models using \texttt{fairseq-py} \cite{ott2019fairseq}\footnote{\url{https://github.com/pytorch/fairseq/}} in PyTorch and train \citet{fan2018hierarchical}'s convolutional architecture. We tune all hyperparameters on validation data. 

\subsection{Generation}
We suppress the generation of unknown tokens to ease human evaluation. For all evaluations, we require stories to be at least 150 words and cut off the story at the nearest sentence for stories longer than 250 words. We generate stories with temperature 0.8 and random top-$k$ sampling method \cameraready{proposed in \cite{fan2018hierarchical}, where next words are sampled from the top $k$ candidates rather than the entire vocabulary distribution. We set $k=10$.}

\section{Experiments}

\subsection{Comparing Decomposition Strategies}

\paragraph{Automated Evaluation} We compare the relative difficulty of modelling using each decomposition strategy by measuring the log loss of the different stages in Table~\ref{tbl:stage_losses}. We observe that generating the SRL structure has a lower negative log-likelihood and so is much easier than generating either summaries, keywords, or compressed sentences --- a benefit of its more structured form. We find keyword generation is especially difficult as the identified keywords are often the more salient, rare words appearing in the story, which are challenging for neural seq2seq models to generate. This result suggests that rare words should appear mostly at the last levels of the decomposition. \cameraready{Further}, we compare models with entity-anonymized stories as an intermediate representation, either with NER-based or coreference-based entity anonymization. Entity references are then filled using a word-based model.\footnote{To make likelihoods are comparable across models.} Perhaps surprisingly, naming entities proves more difficult than creating the entity-anonymized stories---providing insight into the relative difficulty of different sub-problems of story generation.

\cameraready{Finally, we analyze the similarity of the generated stories with the stories in the training set. We quantify this by measuring the maximum and average longest common subsequence of tokens of a generated story with all human-written stories from the training set. High LCS values would indicate models are copying large subparts from existing stories rather than creatively writing new stories. Results shown in Figure~\ref{fig:lcs} indicate that our proposed decomposition copies slightly less long sequences from the training set compared to the baselines --- by separating verb and entity generation into distinct parts, we generate fewer long sequences already present in the training set.}

\begin{figure}
   \centering
   \includegraphics[width=0.5\textwidth]{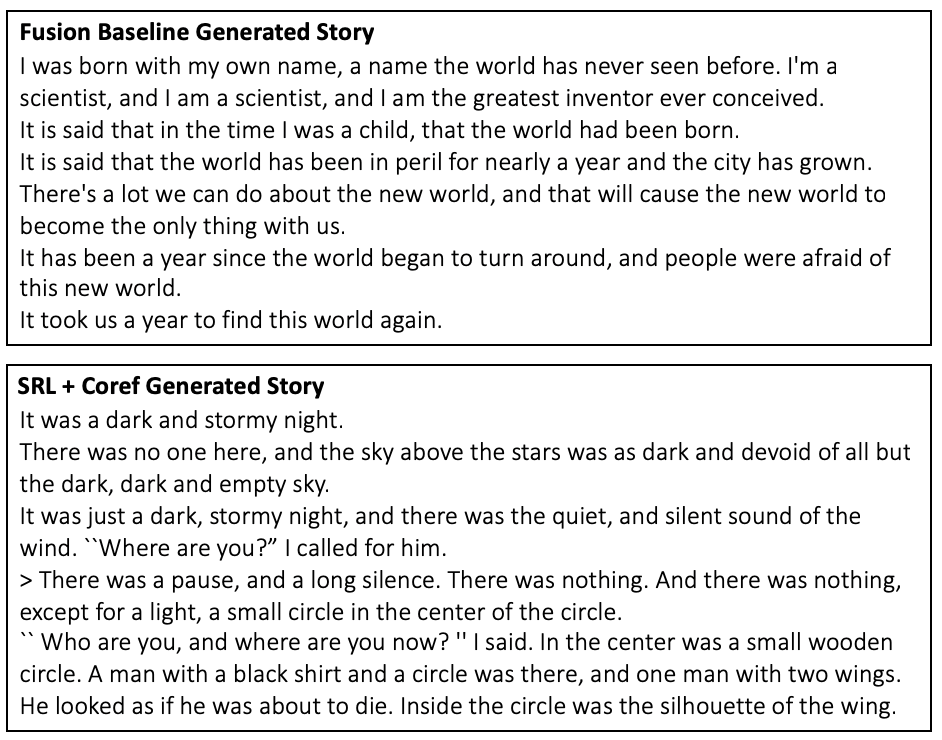}
 \caption{Our decomposition can generate more coherent stories than previous work.}
 \label{fig:example_stories}
\end{figure}

\paragraph{Human Evaluation} To compare overall story quality using various decomposition strategies, we conduct human evaluation \cameraready{using a crowdworking platform}. Judges \cameraready{are shown two different stories that were generated based on the same human-written prompt (but do not see the prompt). Evaluators are asked to mark} which story they prefer. 100 stories are evaluated for each model by 3 \cameraready{different} judges. \cameraready{To reduce variance, stories from all models are trimmed to 200 words.}

Figure~\ref{fig:example_stories} shows that human evaluators prefer our novel decompositions over a carefully tuned Fusion model from \citet{fan2018hierarchical} by about 60\% in a blind comparison. We see additive gains from modelling actions and entities. In a second study, evaluators compared various baselines against stories generated by our strongest model, which uses SRL-based action plans and coreference-based entity anonymization. In all cases, our full decomposition is preferred. 

\subsection{Effect of SRL Decomposition}

Human-written stories feature a wide variety of events, while neural models are plagued by generic generations and repetition. 
Table~\ref{tbl:verb_metrics} quantifies model performance on two metrics to assess action diversity: (1) the number of unique verbs generated, averaged across all stories (2) the percentage of diverse verbs, measured by the percent of all verbs generated in the test set that are not one of the top 5 most frequent verbs. A higher percentage indicates more diverse events.\footnote{We identify verbs using Spacy: \url{https://spacy.io/}}

\begin{figure}
   \centering
   \includegraphics[width=0.5\textwidth]{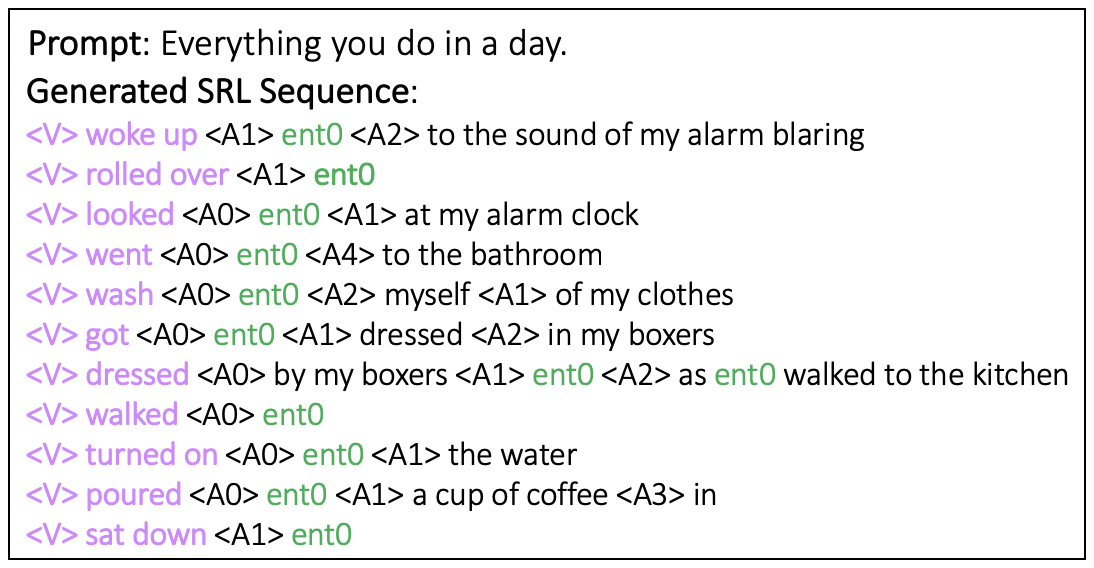}
 \caption{Example \emph{generated} action plan for the SRL + NER Entity Anonymization model. It shows a plausible sequence of actions for a character.}
 \label{fig:srl_example}
\end{figure}

Our decomposition using the SRL predicate-argument structure improves the model's ability to generate diverse verbs. Adding verb attention leads to further improvement. Qualitatively, the model can often outline clear action sequences, as shown in Figure~\ref{fig:srl_example}. However, all models remain far from matching the diversity of human stories.

\begin{table}[t]
  \centering \small
  \begin{tabular}{ l p{13mm} p{14mm} }\hline
    \bf{Model} & \bf{\# Unique Verbs} & \bf{\% Diverse Verbs}\\ \hline\hline
    Human Stories & 34.0 & 76.5 \\ 
    Fusion & 10.3 & 61.1 \\ 
    Summary & 12.4 & 60.6 \\ 
    Keyword & 9.1 & 58.2 \\ 
    Compression & 10.3 & 54.3 \\ 
    \hline  
    SRL & 14.4 & 62.5 \\ 
    + verb-attention & 15.9 & 64.9 \\ 
 \hline 
\end{tabular}
   \caption{Action Generation. Generating the SRL structure improves verb diversity and reduces repetition.}
 \label{tbl:verb_metrics}
\end{table}

\begin{table*}[t]
 \centering \small
  \begin{tabular}{ l p{15mm} p{15mm} p{16mm} | p{15mm} p{15mm} p{16mm}}\hline
    & \multicolumn{3}{c}{\textbf{First Mentions}} & \multicolumn{3}{c}{\textbf{Subsequent Mentions}} \\ 
    \bf{Model} & \bf{Rank 10} & \bf{Rank 50} & \bf{Rank 100} & \bf{Rank 10} & \bf{Rank 50} & \bf{Rank 100}\\ \hline\hline
    Word-Based & 42.3 & 25.4 & 17.2 & 48.1 & 38.4 & 28.8 \\ 
    BPE & 48.1 & 20.3 & 25.5 &  52.5 & 50.7 & 48.8\\ 
    Character-level & 64.2 & 51.0 & 35.6 &  66.1 & 55.0 & 51.2 \\ 
    \hline 
    No story & 50.3 & 40.0 & 26.7 & 54.7 & 51.3 & 30.4\\ 
    Left story context & 59.1 & 49.6 & 33.3 & 62.9 & 53.2 & 49.4\\ 
    Full story & 64.2 & 51.0 & 35.6 & 66.1 & 55.0 & 51.2\\ 
 \hline 
\end{tabular}
   \caption{
   Accuracy at choosing the correct reference string for a mention, discriminating against 10, 50 and 100 random distractors.
   We break out results for the first mention of an entity (requiring novelty to produce an appropriate name in the context) and subsequent references (typically pronouns, nominal references, or shorter forms of names). We compare the effect of sub-word modelling and providing longer contexts.}
 \label{tbl:entity_ranking}
\end{table*}

\begin{table}[t]
  \centering \small
  \begin{tabular}{ l p{23mm}}\hline
    \bf{Model} & \bf{\# Unique Entities}\\ \hline\hline
    Human Stories & 2.99  \\ 
    Fusion & 0.47  \\ 
    Summary & 0.67  \\ 
    Keyword & 0.81  \\ 
    Compression & 0.21  \\ \hline
    SRL + NER Entity Anonymization & 2.16  \\ 
    SRL + Coreference Anonymization & 1.59  \\
 \hline 
\end{tabular}
   \caption{Diversity of entity names. Baseline models generate few unique entities per story. Our decompositions generate more, but still fewer than human stories. 
   Using coreference resolution to build entity clusters reduces diversity here---partly due to re-using existing names more, and partly due to greater use of pronouns.
   }
 \label{tbl:ner_entity_metrics}
\end{table}

\subsection{Comparing Entity Reference Models}

We explored a variety of different ways to generate the full text of abstracted entities---using different amounts of context and different granularities of subword generation. To compare these models, we calculated their accuracy at predicting the correct reference in Table~\ref{tbl:entity_ranking}. Each model evaluates $n=10,  50, 100$ different entities in the test set, 1 real and $n-1$ randomly sampled distractors. Models must give the true mention the highest likelihood. We analyze accuracy on the first mention of an entity, an assessment of novelty, and subsequent references, an assessment of consistency.

\paragraph{Effect of Sub-word Modelling} Table~\ref{tbl:entity_ranking} shows that modelling a character-level vocabulary for entity generation outperforms BPE and word-based models, because of the diversity of entity names. This result highlights a key advantage of multi-stage modelling: the usage of specialized modelling techniques for each sub-task.

\paragraph{Effect of Additional Context} Entity references should be contextual. 
Firstly, names must be appropriate for the story setting---\emph{Bilbo Baggins} might be more appropriate for a fantasy novel.
Subsequent references to the character may be briefer, depending on context---for example, he is more likely to be referred to as \emph{he} or \emph{Bilbo} than his full name in the next sentence.

We compare three models ability to fill entities based on context (using coreference-anonymization): a model that does not receive the story, a model that uses only leftward context (as in \citet{clark2018neural}), and a model with access to the full story. We show in Table~\ref{tbl:entity_ranking} that having access to the full story provides the best performance. Having no access to any of the story decreases ranking accuracy, even though the model still receives the local context window of the entity as input. The left story context model performs better, but looking at the complete story provides additional gains. We note that full-story context can only be provided in a multi-stage generation approach.

\begin{figure}
   \centering
   \includegraphics[width=0.5\textwidth]{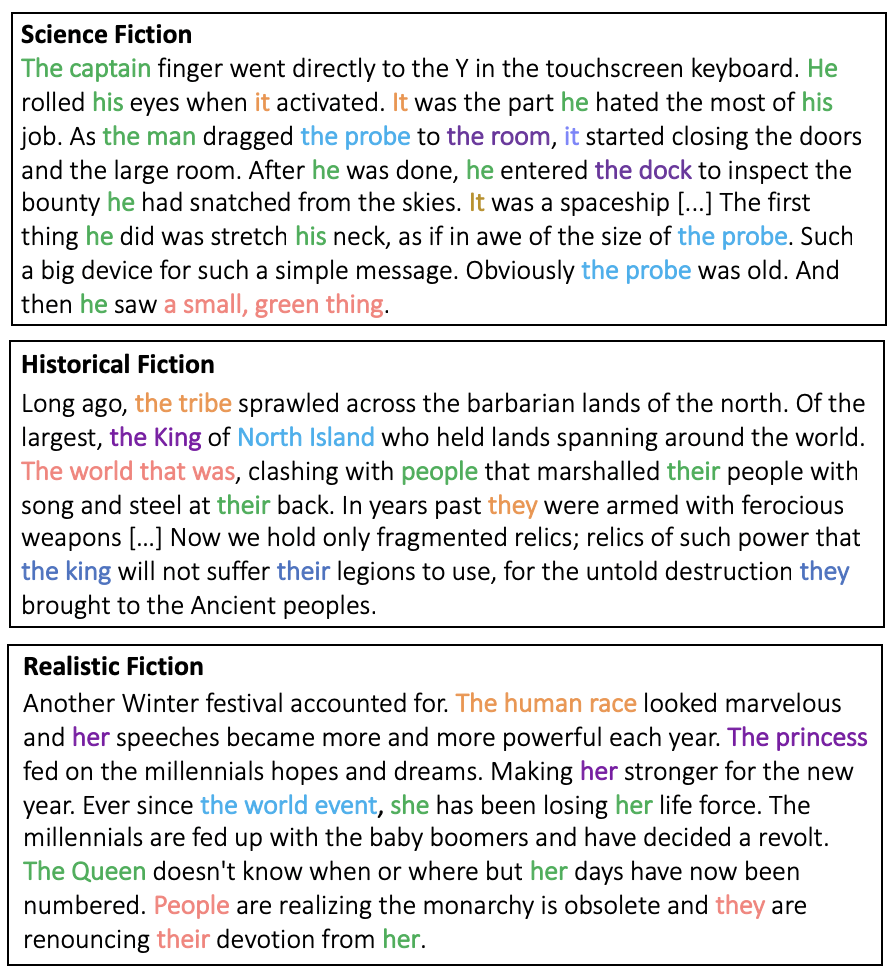}
 \caption{Generating entity references for different genres, using entity-anonymized human written stories. Models use the story context to fill in relevant entities. Color indicates coreferent clusters.}
 \label{fig:ent_fill}
\end{figure}

\paragraph{Qualitative Examples} 
Figure~\ref{fig:ent_fill} shows examples of entity naming in three stories of different genres. \cameraready{We evaluate different genres to examine if generated entities adapt to the style of the story. We show that models can} adapt to the context---for example generating  \textit{The princess} and \textit{The Queen} when the context includes \emph{monarchy}.

\begin{table}[t]
  \centering \small
  \begin{tabular}{ l p{9mm} p{11mm} }\hline
    \bf{Model} & \bf{\# Coref Chains} & \bf{Unique Names per Chain}\\ \hline\hline
    Human Stories & 4.77 & 3.41 \\ 
    Fusion & 2.89 & 2.42 \\ 
    Summary & 3.37 & 2.08 \\ 
    Keyword & 2.34  & 1.65 \\ 
    Compression & 2.84 & 2.09\\\hline 
    SRL + NER Entity Anonymization & 4.09 & 2.49 \\
    SRL + Coreference Anonymization & 4.27 & 3.15 \\ 
 \hline 
\end{tabular}
   \caption{Analysis of non-singleton coreference clusters. Baseline models generate very few different coreference chains, and repetitive mentions within clusters. Our models generate larger and more diverse clusters.}
 \label{tbl:coref_entity_metrics}
\end{table}

\subsection{Effect of Entity Anonymization} 

To understand the effectiveness of the entity generation models, we examine their performance by analyzing generation diversity.

\paragraph{Diversity of Entity Names} 
Human-written stories often contain many diverse, novel names for people and places. However, these tokens are rare and subsequently difficult for standard neural models to generate. Table~\ref{tbl:ner_entity_metrics} shows that the fusion model and baseline decomposition strategies generate very few unique entities in each story. Generated entities are often generic names such as \textit{John}. 

Our proposed decompositions generate substantially more unique entities than strong baselines. Interestingly, we found that using coreference resolution for entity anonymization led to fewer unique entity names than generating the names independently. This result can be explained by the coreference-based model re-using previous names more frequently, as well as using more pronouns. 

\paragraph{Coherence of Entity Clusters} 
Well structured stories will refer back to previously mentioned characters and events in a consistent manner. To evaluate if the generated stories have these characteristics, we examine the coreference properties in Table~\ref{tbl:coref_entity_metrics}. We quantify the average number of coreference clusters and the diversity of entities within each cluster (e.g. the cluster \textit{Bilbo, he, the hobbit} is more diverse than the cluster \textit{he, he, he}). 

Our full model produces more non-singleton coreference chains, suggesting greater coherence, and also gives different mentions of the same entity more diverse names. However, both numbers are still lower than for human generated stories, indicating potential for future work.

\begin{figure}[t]
   \centering
   \includegraphics[width=0.5\textwidth]{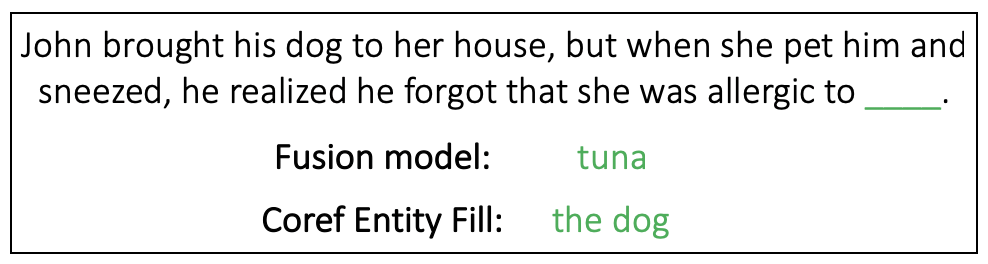}
 \caption{Constructed sentence where the last word refers to an entity. The coreference model is able to track the entities, whereas the fusion model relies heavily on local context to generate the next words.}
 \label{fig:winograd}
\end{figure}

\paragraph{Qualitative Example} Figure~\ref{fig:winograd} displays a sentence constructed to require the generation of an entity as the final word. The fusion model does not perform any implicit coreference to associate the \textit{allergy} with \textit{his dog}. In contrast, coreference entity fill produces a high quality completion.

\section{Related Work}

\cameraready{Decomposing natural language generation into several steps has been extensively explored \cite{reiter2000building,gatt2018survey}. In classical approaches to text generation, various stages were used to produce final written text. For example, algorithms were developed to determine content and discourse at an abstract level, then sentence aggregation and lexicalization, and finally steps to resolve referring expressions \cite{hovy1990pragmatics,dalianis1993aggregation,wahlster1993plan,ratnaparkhi2000trainable,malouf2000order}. Our work builds upon these approaches.}

\paragraph{Story Generation with Planning}

Story generation using a plan has been explored using many different techniques. Traditional approaches organized sequences of character actions with hand crafted models \cite{riedl2010narrative, porteous2009controlling}. Recent work  extended this to modelling story events \cite{martin2017event, mostafazadeh2016caters}, plot graphs \cite{li2013story}, \cameraready{plot summaries \cite{appling2009representations}, story fragments or vignettes \cite{riedl2010story}}, or used sequences of images \cite{huang2016visual} or descriptions \cite{jain2017story}. 

We build on previous work that decomposes generation. \citet{xu2018skeleton} learn a skeleton extraction model and a generative model conditioned upon the skeleton, using reinforcement learning to train jointly. 
\citet{zhou2018neural} train a storyline extraction model for news articles, but require supervision from manually annotated storylines. \citet{yao2019plan} use  \textsc{rake} \cite{rose2010automatic} to extract storylines, and condition upon the storyline to write the story using dynamic and static schemas that govern if the storyline can change. 

\paragraph{Entity Language Models}

An outstanding challenge in text generation is modelling and tracking entities. 
Centering \citep{grosz1995centering} gives a theoretical account of how referring expressions for entities are chosen in discourse context.
Named entity recognition has been incorporated into language models since at least \citet{gotoh1999named}, and can improve domain adaptation \cite{liu2007unsupervised}. Language models have been extended to model entities based on information such as entity type \cite{parvez2018building}. Recent work has incorporated learning representations of entities and other unknown words \cite{kobayashi2017neural}, as well as explicitly model entities by dynamically updating these representations to track changes over time and context \cite{ji2017dynamic}. Dynamic updates to entity representations are used in other story generation models \cite{clark2018neural}. 

\paragraph{Non-Autoregressive Generation}
\cameraready{Our method proposes decomposing left-to-right generation into multiple steps. Recent work has explored non-autoregressive generation for more efficient language modeling and machine translation.} \citet{ford2018importance} developed two-pass language models, generating templates then filling in words. \cameraready{The partially filled templates could be seen as an intermediary representation similar to generating a compressed story.} Other models allow arbitrary order generation using insertion operations \cite{gu2019insertion, stern2019insertion} and \citet{gu2017non} explored parallel decoding for machine translation. \cameraready{In contrast, we focus on decomposing generation to focus on planning, rather than efficient decoding at inference time.}

\section{Conclusion} 

We proposed an effective method for writing short stories by separating the generation of actions and entities. We show through human evaluation and automated metrics that our novel decomposition improves story quality. 

\bibliography{acl2019}
\bibliographystyle{acl_natbib}

\end{document}